\pdfoutput=1
\documentclass{article}
\usepackage{microtype}
\usepackage{booktabs} 
\usepackage{amsmath}
\usepackage{amssymb}
\usepackage{url}
\usepackage{graphicx,subfigure}
\graphicspath{ {./figures/} }

\usepackage[accepted]{icml2019}

\icmltitlerunning{Venn GAN}

\begin{document}

\twocolumn[
\icmltitle{Venn GAN: \\
Discovering Commonalities and Particularities of Multiple Distributions}



\icmlsetsymbol{equal}{*}

\begin{icmlauthorlist}
\icmlauthor{Yasin Yaz{\i}c{\i}}{ntu}
\icmlauthor{Bruno Lecouat}{i2r}
\icmlauthor{Chuan-Sheng Foo}{i2r}
\icmlauthor{Stefan Winkler}{nus}
\icmlauthor{Kim-Hui Yap}{ntu}
\icmlauthor{Georgios Piliouras}{sutd}
\icmlauthor{Vijay Chandrasekhar}{i2r}
\end{icmlauthorlist}

\icmlaffiliation{ntu}{School of Electrical and Electronics Engineering, Nanyang Technological University, Singapore}
\icmlaffiliation{i2r}{Institute for Infocomm Research, A*STAR, Singapore}
\icmlaffiliation{nus}{School of Computing, National University of Singapore, Singapore}
\icmlaffiliation{sutd}{Engineering Systems and Design, Singapore University of Technology and Design, Singapore}

\icmlcorrespondingauthor{Yasin Yaz{\i}c{\i}}{yasin001@e.ntu.edu.sg}

\icmlkeywords{Generative Adversarial Networks (GANs), Generative Models}

\vskip 0.3in
]



\printAffiliationsAndNotice{}  

\begin{abstract}
We propose a GAN design which models multiple distributions effectively and discovers their commonalities and particularities. Each data distribution is modeled with a mixture of $K$ generator distributions. As the generators are partially shared between the modeling of different true data distributions, shared ones captures the commonality of the distributions, while non-shared ones capture unique aspects of them. We show the effectiveness of our method on various datasets  (MNIST, Fashion MNIST, CIFAR-10, Omniglot, CelebA) with compelling results\footnote{The code can be found here: \url{https://github.com/yasinyazici/Venn_GAN}}.
\end{abstract}

\section{Introduction}

Generative Adversarial Networks (GAN) \cite{gan} learn a function that can sample from an approximated probability distribution. Due to enormous interest, GAN have been improved substantially over the past few years \cite{dcgan, wgan_gp, spectral_norm, pggan, convergence_lars}. 

\begin{figure}[h!]
	\centering
	\includegraphics[width=1.0\linewidth]{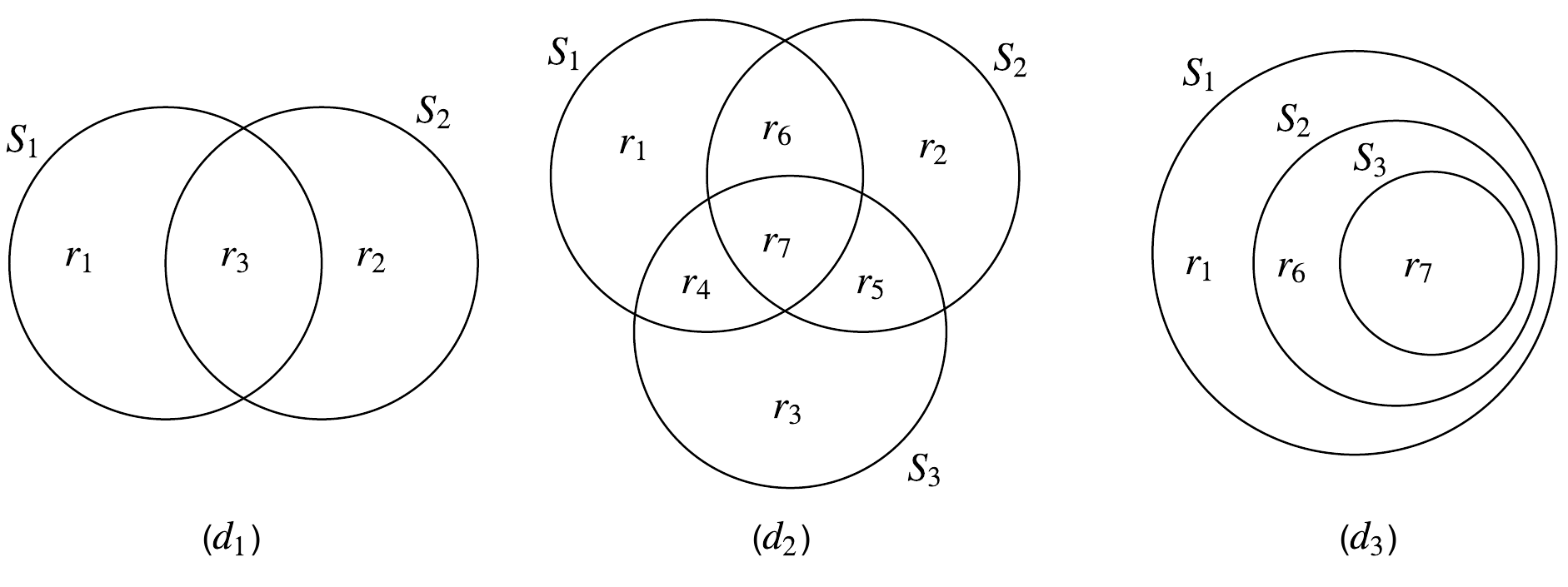}
	\caption{Three different configurations of Venn diagrams with 2 and 3 sets.}
	\label{venn_diagrams}
\end{figure}

GANs are designed to learn a single distribution, though multiple distributions can be modeled by treating them separately. However, this naive implementation does not consider relationships between the distributions. An interesting question is how we can model multiple distributions efficiently and discover their common and unique aspects? We explain this situation by utilizing Venn diagrams. Figure \ref{venn_diagrams} depicts some cases of different interactions between 3 sets, where each set represents a distribution. In $d_2$, each set has its own unique part and intersections with the other sets, whereas in $d_3$, some sets are a \textit{superset} of others. Each case can be useful in different scenarios, e.g.\ $d_3$ can be used in a case where a distribution is a subset of another distribution, such as a specific dog breed and its \textit{superset} is many different dog-breeds.

In this paper, we propose Venn~GAN, which models multiple distributions efficiently and discovers their interactions and uniqueness. Each data distribution is modeled with a mixture of $K$ generator distributions. As the generators are partially shared between the modeling of different true data distributions, shared ones captures the commonality of the distributions, while non-shared ones capture unique aspects of them. Our contributions are the following:
\begin{itemize}
    \item Introducing a novel and interesting problem setting where there exists multiple distribution various configurations (See Figure~\ref{venn_diagrams}).
    \item Proposing a new method that can capture commonalities and particularities of various distributions with high success rate.
    \item Thoroughly evaluating the method on various datasets, namely MNIST, Fashion MNIST, CIFAR-10, Omniglot, CelebA, with compelling results.
\end{itemize}

\section{Related work}
\textbf{Multi-generator/discriminator GAN:} There have been some attempts to use multiple generator/discriminator in order to solve various issues with GAN. \citet{DBLP:journals/corr/Arora0LMZ17,mggan,madgan} modeled a single distribution with multiple generators to capture different modes of the distribution. In order to guide the generators into different modes, they utilized a classifier which separates each generator from one another. \citet{multi_disc_ishan, DBLP:journals/corr/NeyshaburBC17, DBLP:journals/corr/Juefei-XuBS17} utilized multiple discriminators to address mode collapse and optimization stability. Similarly, \citet{gan_multi_armed_bandit} used multiple discriminators with learned importance to ease training of GAN. \citet{adaGAN} used a meta-learning algorithm analogous to \textit{AdaBoost} to improve coverage of modes with multiple generators.

\textbf{Mixture of Distributions with GAN:} Some of the earlier works considered multiple generators as mixture of distributions to model a single distribution \cite{DBLP:journals/corr/Arora0LMZ17,mggan,madgan}. Our model is different, as we model multiple data distributions and share the generator distributions as component for each data distribution.

\textbf{Conditional GAN:} This type of GAN uses a condition, alongside noise, to generate data \cite{DBLP:journals/corr/MirzaO14}. The conditions are desired to correlate with generated data. It has been used for Image-to-Image transformation \cite{DBLP:journals/corr/IsolaZZE16, hoang2018mgan, DBLP:journals/corr/YiZTG17}, text-to-image \cite{DBLP:journals/corr/ReedAYLSL16}, super resolution \cite{DBLP:journals/corr/LedigTHCATTWS16}.
 
The way GANs are conditioned is still an active research field. We have focused on conditioning of the generator. The most common way to include conditions into the generator is to provide it as input \cite{DBLP:journals/corr/MirzaO14,DBLP:journals/corr/ReedAYLSL16,2016arXiv161009585O}. Recently, \citet{miyato2018cgans} used conditional BatchNorm \cite{devries+al-2017-modulating, dumoulin2017learned-iclr} to include conditions into generator.


\textbf{Other Related Works:} Concurrent work of \cite{DBLP:journals/corr/abs-1811-11163} is perhaps the most similar work to ours. However, their motivation, method and experiments are different then ours. They are motivated by ambiguous class labels, due to noisy labels, and propose a model to discover class-distinct and class-mutual parts. Their method utilizes modified version of AC-GAN and redesigns input of $G$ to achieve the objective. While our work scales GAN objective into $n$ distributions and models each distribution as mixture of generator distributions.

\section{Method}
\subsection{Background}

GAN is a two player zero-sum game between a discriminator and generator:
\begin{equation}\label{minimax_objective_one}
    \min_{G} \max_{D} V(D,G)
\end{equation}
\begin{align}\label{minimax_objective_two}
 V(D,G) = E_{x \sim p_{data}(\boldsymbol x)}[\log D(\boldsymbol x)] + \\
 E_{x \sim p_{g}(\boldsymbol x)}[\log(1-D(\boldsymbol x))] \nonumber
\end{align}

It utilizes a discriminator to assess a peudo-divergence between the true data distribution, $p_{data}(\boldsymbol x)$, and the generator's distribution, $p_{g}(\boldsymbol x)$. The discriminator maximizes the divergence, while the generator minimizes it. In this way, the generator learns to mimic the data distribution implicitly. \citet{gan} show that, under certain assumptions, for a fixed optimal $D$, minimizing Eq.~\ref{minimax_objective_two} for $G$ would lead to $p_{g}(\boldsymbol x) = p_{data}(\boldsymbol x)$.

\subsection{Multi-distribution GAN}

The value function, Eq.~\ref{minimax_objective_one}, can be scaled to $n$ distributions trivially as follows:
\begin{equation}\label{minimax_objective_multi_one}
    \min_{G_1,...,G_n} \max_{D_1,...,D_n} V(D_1,D_2,...,D_n,G_1,G_2,...,G_n)
\end{equation}
\begin{equation}
\begin{aligned}\label{minimax_objective_multi_two}
    V(D_1,D_2,...,D_n,G_1,G_2,...,G_n) = \\ 
    \frac{1}{n}\sum_{i=1}^n E_{x \sim p_{data_i}(\boldsymbol x)}[\log D_i(\boldsymbol x)] + \\
    \frac{1}{n}\sum_{i=1}^n E_{x \sim p_{g_i}(\boldsymbol x)}[\log(1-D_i(\boldsymbol x))]
\end{aligned}
\end{equation}
where $p_{data_i}(\boldsymbol x)$ is $i$-th true data distribution and $p_{g_i}(\boldsymbol x)$ is $i$-th generator's distribution, which are independent from one another. Note that $D_i$ and $G_j$\footnote{Eq.~\ref{minimax_objective_multi_two} does not explicitly show $G_j$ but $p_{g_j}$ which is distribution of $j$-th generator, $G_j$} in above equation interact with one another only when $i=j$. This makes learning one distribution independent from the others. By following the proof from \citet{gan}, we can show that, at equilibrium $\boldsymbol{p_{data}} = \boldsymbol{p_g}$.

However this objective does not consider possible overlaps between the data distributions. Incorporating this can make the model more efficient and leads to interesting discoveries, e.g.\ commonalities and particularities of the distributions. In order to achieve this, we have reformulated the way we construct generator distributions, $p_{g_i}$. It is no longer equal to the distribution of $i$-th generator, but a mixture of $K$ generator distributions, denoted by $p_{r_i}$, Eq.~\ref{eq:mixture_dist}. In this way, each data distribution is modeled as a mixture of generators' distributions. As $p_{r_i}$ are shared for all data distributions, some of them cover common parts and others unique ones. Each generators learns only sub-part of the distributions and combines them at different amounts to make the data distributions.

\begin{equation}\label{eq:mixture_dist}
    \begin{bmatrix}
        p_{g_1} \\
        p_{g_2} \\
        \vdots \\
        p_{g_{n}}
    \end{bmatrix}
    =
    \begin{bmatrix}
        o_{11} & o_{12} & \cdots & o_{1K} \\
        o_{21} & o_{22} & \cdots & o_{2K} \\
        \vdots & \vdots & \ddots & \vdots \\
        o_{n1} & o_{n2} & \cdots & o_{nK}
    \end{bmatrix}
    \begin{bmatrix}
        p_{r_1} \\
        p_{r_2} \\
        \vdots \\
        p_{r_{K}}
    \end{bmatrix}
\end{equation}
where $\boldsymbol{O} \in \mathbb{R}^{n \times K}$ is a mixture matrix whose rows sum up to one to make $p_{g_i}$ valid. Note that this reformulation does not change the objective (Eq.~\ref{minimax_objective_multi_one} and Eq.~\ref{minimax_objective_multi_two}), but how we model $p_{g_i}$.

\subsection{Conceptual Explanation: Relation to Venn Diagrams}\label{venn_gan_explanation}

The method in the previous section can be explained by using Venn diagrams where each set represents a distribution. We deal with a situation where multiple distributions exist. Each distribution might have a unique part and commonalities with other distributions e.g. $d_2$ of Figure \ref{venn_diagrams}. In another case, one distribution's support might cover the others' e.g. $d_3$ of Figure \ref{venn_diagrams}. Our proposed method models each region of a Venn diagrams as a probability distribution $p_{r_i}(x)$. Each set should capture the distribution of its corresponding data distribution, e.g. $p_{S_i} = p_{g_i} = p_{data_i}$. Each set can be represented by union of its regions, e.g. $d_2$ of Figure \ref{venn_diagrams}, $S_1 = r_1 \cup r_4 \cup r_6 \cup r_7$. Similarly, each region can be represented with set operations e.g. $d_3$ of Figure \ref{venn_diagrams}, $r_1 = S_1 \setminus (S_1 \cup S_3) $. Set configurations can be in different forms e.g. $d_3$ of Figure \ref{venn_diagrams} is $S_3 \subset S_2 \subset S_1$. 
 
$d_3$ type diagram can be represented by:
\begin{equation}
    \boldsymbol{O}
    = 
    \begin{bmatrix}
        \frac{1}{3} & 0 & 0 & 0 & 0 & \frac{1}{3} & \frac{1}{3} \\
        0 & 0 & 0 & 0 & 0 & \frac{1}{2} & \frac{1}{2} \\
        0 & 0 & 0 & 0 & 0 & 0 & 1
    \end{bmatrix}
\end{equation}

Similarly $d_2$ type diagram can be represented by:
\begin{equation}
    \boldsymbol{O}
    = 
    \begin{bmatrix}
        \frac{1}{4} & 0 & 0 & \frac{1}{4} & 0 & \frac{1}{4} & \frac{1}{4} \\
        0 & \frac{1}{4} & 0 & 0 & \frac{1}{4} & \frac{1}{4} & \frac{1}{4} \\
        0 & 0 & \frac{1}{4} & \frac{1}{4} & \frac{1}{4} & 0 & \frac{1}{4}
    \end{bmatrix}
\end{equation}

In both cases we assume that each region contributes equally. Learning mixture weights is left for future study.

\subsection{Implementation Details}
\textbf{Generator side:} We can use two approaches to model the generators ($G_1, G_2, ..., G_K$). The first is the use of $K$ independent generators for each region. Each generator is modeled by $r_i = G_{i}(z;\theta_i)$, where $G$ is a generative network, $z$ is input noise and $\theta_i$ are the parameters of the $i$-th network. The second approach is a single generator with $K$ conditions. Each region is modeled with a function $r_i = G(z,c=i;\theta)$, $c$ is a condition whose $i$-th index used to generate region $r_i$ and $\theta$ are the network parameters. The former approach can be expensive when there are many region to model, however it has its own advantage as we will show in the experiments. Conditional generator is more efficient as the number of regions grows exponentially with distributions, e.g.\ $n$ distributions contain up to $2^n -1$ regions. Also, sharing weights with other generators regularizes the model and makes the training easier. Besides, using this type of generator has certain effects on modeling, namely different conditions with the same noise produce semantically related samples, as detailed in the CelebA experiments. We use both types and discuss their advantages and disadvantages in more detail in the experiment section.

\begin{figure}[ht!]
	\centering
	\includegraphics[width=1.0\linewidth]{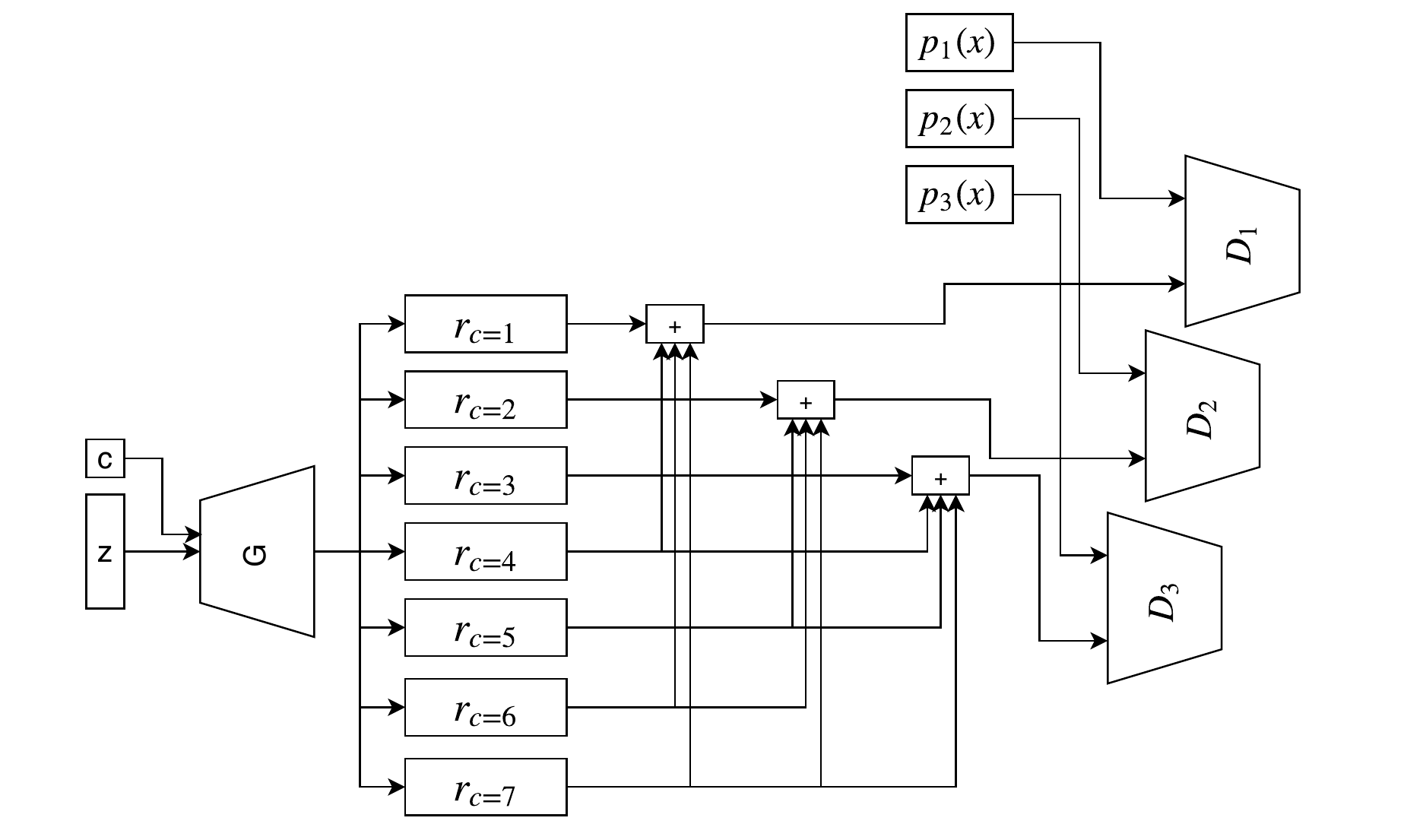}
	\caption{Venn~GAN architecture for 3 distributions and $d_2$ type diagrams. Each mode of the generator represents corresponding region in Figure \ref{venn_diagrams}; other Venn diagrams can be constructed in a similar way. Each discriminator receives union of its corresponding set's regions. For this illustration we have used conditional generator, however $K$ independent generators can be also used in the same way. ``$+$'' sign takes union over its incoming regions, and the union is feed into a discriminator as fake class. Each discriminator receives different true data distribution and with GAN objective it compares them to mixture of the regions.}
	\label{venn_gan}
\end{figure}

\textbf{Discriminator side:} There should be $n$ discriminators for $n$-distribution game. As we have changed generator distribution into a mixture of distributions, each discriminator takes input from all incoming generators, which has non-zero mixture weight. Figure \ref{venn_gan} illustrates how a $d_2$ type diagrams looks like in terms on connections. Other types can be constructed in a similar way by following the connection pattern from the weight matrix $\boldsymbol{O}$. When sampling from the generators to feed into $D_i$, the number of samples from each generator should be proportional to $i$-th row of $\boldsymbol{O}$. The ``$+$'' sign in the diagram corresponds to union operation over the incoming regions. In practice it is concatenation over batch dimensions. As each set should represent a true data distribution, $p_{S_i}(x) = p_{data_i}(x)$, union of regions that belongs to $S_i$ should match to $i-th$ data distribution. In order to satisfy this, each discriminator, $D_i$, compares a specific true data distribution, $p_{data_i}(x)$, with union of regions, $p_{S_i}(x)$, which belongs to the corresponding set e.g. $S_1 = r_1 \cup r_4 \cup r_6 \cup r_7$. As certain regions are fed into more than one discriminator, those regions would be forced to represent common parts of the distributions. For example, $r_7$ will suffer a loss if its modeling does not satisfy the 3-way intersection of the distributions. In other words, it will receive a negative feedback from the discriminator(s) which it could not satisfy. Similar analogies can be made to $r_4$, $r_5$, $r_7$ which are 2-way intersections, whereas individual regions like $r_1$, $r_2$, $r_3$ are only used by a single discriminator, thereby they are inclined to model the unique part of its corresponding distribution. Sharing the regions between different discriminators which receive different true data distributions is the core dynamic of learning commonalities between true data distributions. We make the assumption that all the regions in a distribution have equal weights. 

The objective of the model is a minimax game with $n$ discriminators for $n$-distribution game is stated in Eq.~\ref{minimax_objective_multi_one} and Eq.~\ref{minimax_objective_multi_two}. In Eq.~\ref{eq:mixture_dist}, we show how $p_{g_i}$ can be represented. From Venn diagrams perspective, it can be also represented by:
\begin{equation}
    p_{g_i}(\boldsymbol x) = \frac{1}{|S_i|} \sum_{r_j \in S_i} p_{r_j}
\end{equation}
where $|S_i|$ is number of regions in set $S_i$.

In practice we observe that there is some amount of leakage between regions. In order to alleviate this issue, we include an additional objective, which aims to separate regions of the generator from one another:
\begin{equation}
\begin{split}
    \max_{C,G_1,G_2,...,G_n} E_{z \sim p_{z}(\boldsymbol z)} \log C(y_i | \boldsymbol r_i ; \phi_c)\\
    = E_{z \sim p_{z}(\boldsymbol z)} \log C(y_i | G_i(z; \theta_i) ; \phi_c)
\end{split}
\label{classification_formulation}
\end{equation}
where $y_i$ is the category for $\boldsymbol r_i$ and $C$ is a classifier which outputs probability distribution over the regions. With this objective, the classifier tries to separate the regions and the generator tries to satisfy the classifier by increasing differences between the regions. Similar losses has been used by \cite{hoang2018mgan} previously. The combined objective becomes:
\begin{equation}
\begin{split}
    \min_{C,G_1,G_2,...,G_n} \max_{D_1,D_2, ..., D_n} V(D_1,D_2, ..., D_n,G_1,G_2,...,G_n) \\
    - \lambda E_{z \sim p_{z}(\boldsymbol z)} \log C(y_i | G_i(z))
\end{split}
\label{overall_equation}
\end{equation}
where $\lambda$ is balancing hyper-parameter between the two terms.

\section{Experiments} 

\textbf{Network Architecture:} Discriminator and generator architectures are similar to DCGAN \cite{dcgan} for MNIST, Fashion-MNIST, Omniglot and CIFAR-10, while CelebA uses ResNet type architecture with detailed specifications given in the Appendix. The classifier architecture is the same as the discriminator except for the last layer, whose output dimensions equal the number of regions. Exponential Moving Average (EMA) \cite{pggan,yasin_ema} has been used over generator(s) parameters out of training loop. Conditioning of $G$ is similar to that of \citet{miyato2018cgans,devries+al-2017-modulating,dumoulin2017learned-iclr} except that there is no normalization but scaling and addition.



\textbf{Objective Details:} Zero gradient penalty \cite{convergence_lars} has been applied on true data distributions for each discriminator with weight $1.0$ in every case but illustrative examples. We found that this improves the quality of generation, especially in CelebA.

\textbf{Optimization \& Hyperparameters:} We have used ADAM \cite{adam} optimizer with learning rate of $0.0002$, $\beta_1 = 0.0$ and $\beta_2 =0.9$. The optimization of discriminator and generator follows alternating update rule with single discriminator update per generator update. The model has been trained for 100k iterations for CelebA, 50k for CIFAR-10, 20k for MNIST, Fashion-MNIST and Omniglot. For each region, we use a batch size of $16$, except for illustrative example which uses $64$. The batch size of real data depends on the number of regions fed to each discriminator. Union over $k$ regions would corresponds to a batch size of $16k$. $\lambda$ is selected as $0.1$ by searching over range of $[0.1, 10.0]$ with quantitative score (will be explained shortly) over various scenarios. Classifier's optimization is the same with the discriminators'. For the conditional generator, we have used the same noise for different conditions during training. The illustrative example does not use a classifier.

\textbf{Quantification of Results:} In case of artificial datasets, we can quantify the rate of correct generation (accuracy) for different regions. In order to achieve this, we have trained a separate classifier on MNIST, fashion-MNIST \cite{fashion_mnist} and CIFAR-10 \cite{cifar10} by using their training data split. This model is used to assess if the generated images from each region belongs to the correct class. We use 10k generated samples from each region to assess the quantity. The accuracy of the classifier on each region is used as the metric. The details about architecture, optimization etc.\ for the classifier can be found in the Appendix. The accuracy of the classifier on test sets for MNIST, fashion-MNIST and CIFAR-10 are 99.12, 91.20 and 84.20 respectively. During the VennGAN training, we have measured the model at every 2k iterations and report the best average results.

\subsection{Illustrative Examples}
\begin{figure}[ht!]
	\centering
    \includegraphics[width=1.0\linewidth]{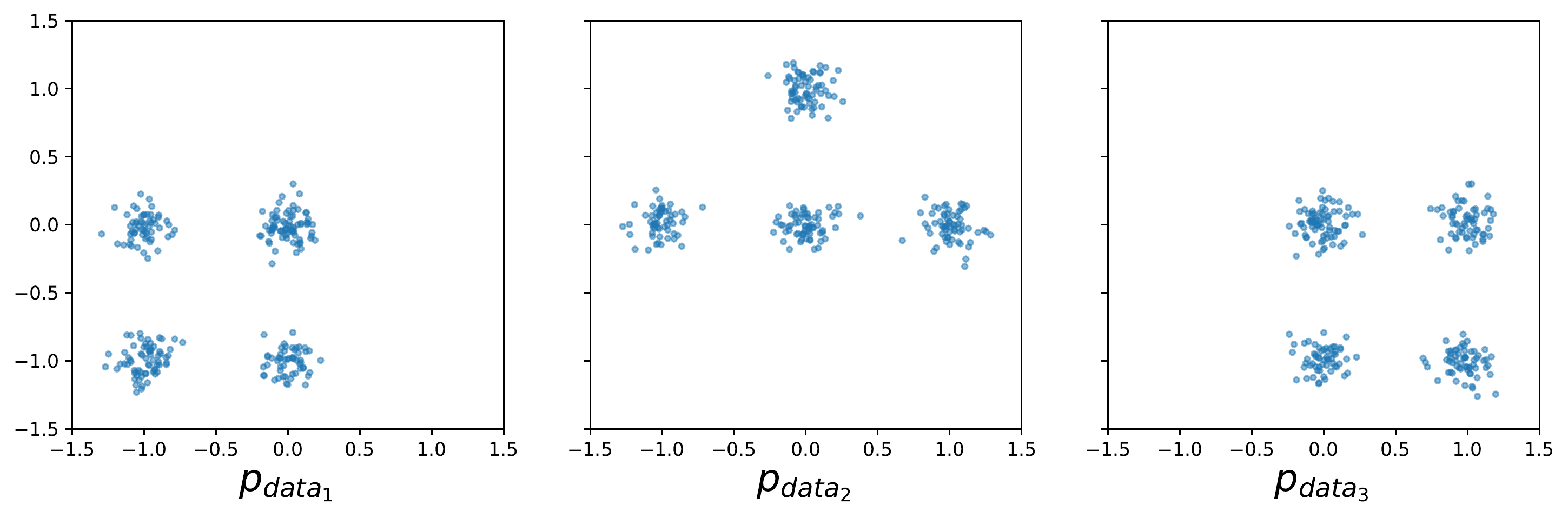}
    \vspace{-4mm}
	\caption{Samples from the data distributions for the illustrative example. Each distributions equally mixes 4 out of 7 Gaussians.}
	\label{fig:gaussian_toy}
\end{figure}

We use mixture of Gaussians illustrative example to show that the method works as anticipated. The nature of the dataset and its dimensionality make it easier to spot subtle behaviours of the method. For this experiment we generate 3 different data distributions where each data distribution equally mixes 4 out of 7 Gaussians as in Figure~\ref{fig:gaussian_toy}.

In order to model these distributions, we have used $d_2$ type with $n=3$. The experiment is conducted with independent generators for 5k iterations. Further details about the training, architecture etc.\ are in Appendix. Figure~\ref{fig:gaussian_toy_results} shows the results. All the regions are generated at the correct position, e.g.\ the pink samples generated by $r_7$, which is the common mode of all the distributions. We have conducted this experiment multiple times with no notable differences which shows stability of the method.

\begin{figure}[ht!]
	\centering
    \includegraphics[width=0.7\linewidth]{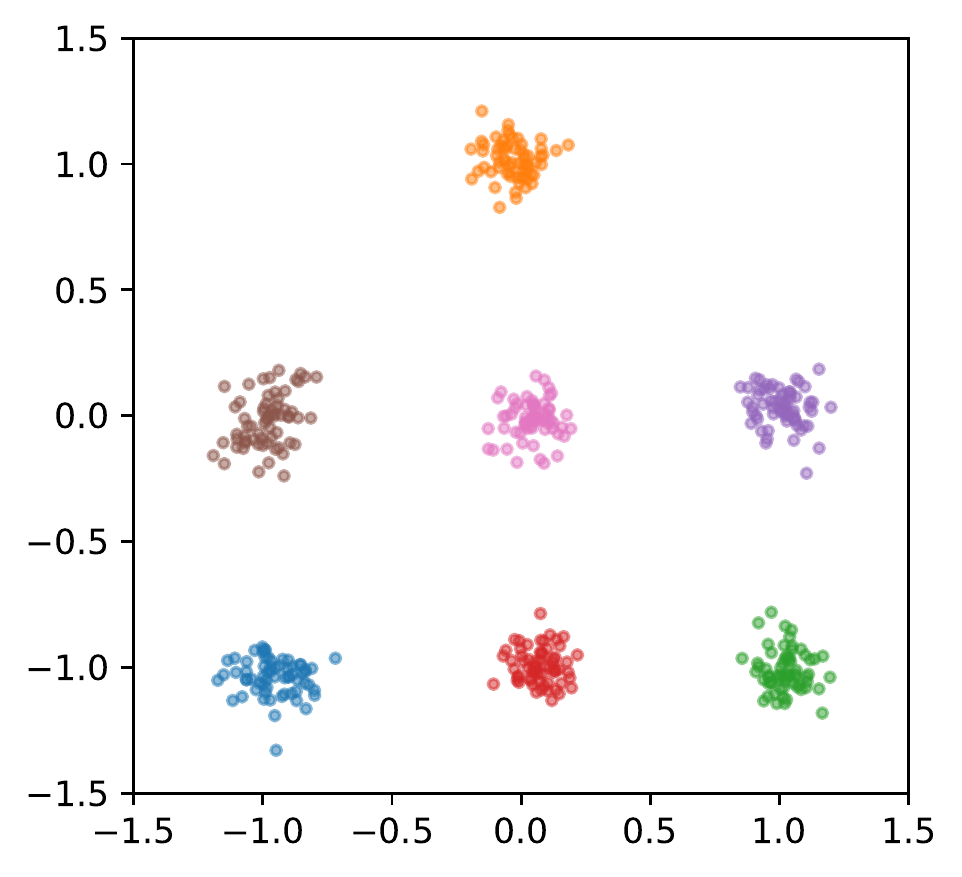}
    \vspace{-4mm}
	\caption{Generated regions for the illustrative example. Annotation of the regions w.r.t.\ $d_2$ of Figure~\ref{venn_diagrams}: $r_1$ is blue, $r_2$ is orange, $r_3$ is green, $r_4$ is brown, $r_5$ is purple, $r_6$ is red and $r_7$ is pink.}
	\label{fig:gaussian_toy_results}
\end{figure}

\subsection{Main Experiments}

We have designed multiple artificial datasets as well as natural datasets to investigate the working dynamic of the method. For artificial datasets, MNIST, fashion-MNIST \cite{fashion_mnist} and CIFAR-10 \cite{cifar10} have been used. By using these datasets, we have designed 2 and 3 distribution games with $d_1$, $d_2$ and $d_3$ type Venn diagrams. The distributions are constructed by using the class information from the datasets as per Table \ref{dataset_configuration}. For all types, each distribution contains 2000 samples from the classes it includes. We never use the same sample twice for different distributions, which could lead to trivial solutions.

\begin{table*}[ht!]
\caption{Configurations of artificial datasets}
\resizebox{\linewidth}{!}{\begin{tabular}{cccc}
 \hline
 Case & Venn Type & Distributions & Sets  \\
 \hline
 A & $d_1$ & 2 & $S_1 = \left\{ {0,1,2,3,4,5,6}\right\}$, $S_2 = \left\{ {3,4,5,6,7,8,9}\right\}$ \\
 B & $d_2$ & 3 &  $S_1 = \left\{ {0,3,5,6}\right\}$, $S_2 = \left\{ {1,4,5,6}\right\}$, $S_3 = \left\{ {2,3,4,6}\right\}$ \\
 C & $d_3$ & 3 & $S_1 = \left\{ {0,1,2,3,4,5,6,7,8,9}\right\}$, $S_2 = \left\{ {3,4,5,6,7,8,9}\right\}$, $S_3 = \left\{ {6,7,8,9}\right\}$  \\
 \hline
\end{tabular}}
\label{dataset_configuration}
\end{table*}

\begin{table*}[ht!]
\caption{Correspondence of labels for fashion-MNIST and CIFAR-10}
\resizebox{\linewidth}{!}{\begin{tabular}{ccccccccccc}
 \hline
 \emph{} & $0$ & $1$ & $2$ & $3$ & $4$ & $5$ & $6$ & $7$ & $8$ & $9$ \\
 \hline
 Fashion-MNIST & T-shirt/top & Trouser & Pullover & Dress & Coat & Sandal & Shirt & Sneaker & Bag & Ankle boot \\
 Cifar-10 & Airplane & Automobile &  Bird & Cat & Deer &  Dog & Frog & Horse & Ship & Truck \\
 \hline
\end{tabular}}
\label{dataset_configuration}
\end{table*}

\begin{figure}[ht!]
	\centering
    \subfigure{\includegraphics[width=1.0\columnwidth]{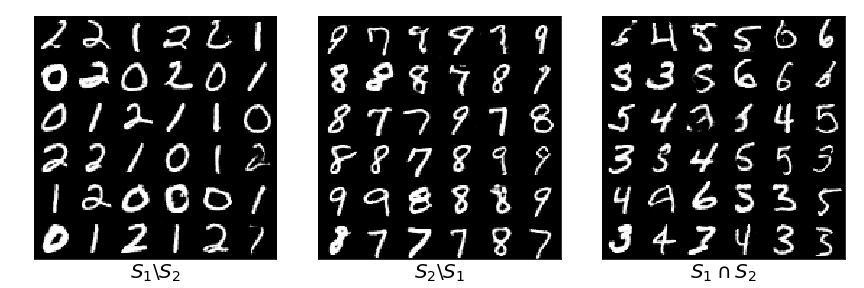}}
    \subfigure{\includegraphics[width=1.0\columnwidth]{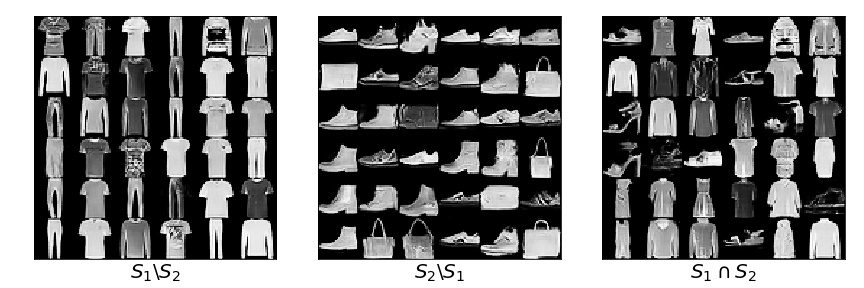}}
    \subfigure{\includegraphics[width=1.0\columnwidth]{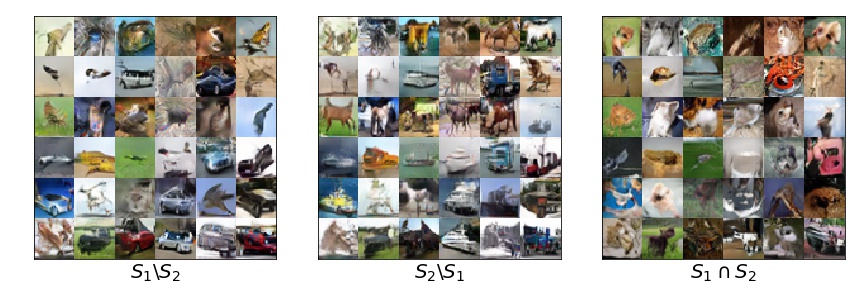}}
    \vspace{-4mm}
	\caption{MNIST, Fashion-MNIST, CIFAR-10 results for case A}
	\label{case1}
\end{figure}

\begin{figure}[ht!]
	\centering
    \subfigure{\includegraphics[width=1.0\linewidth]{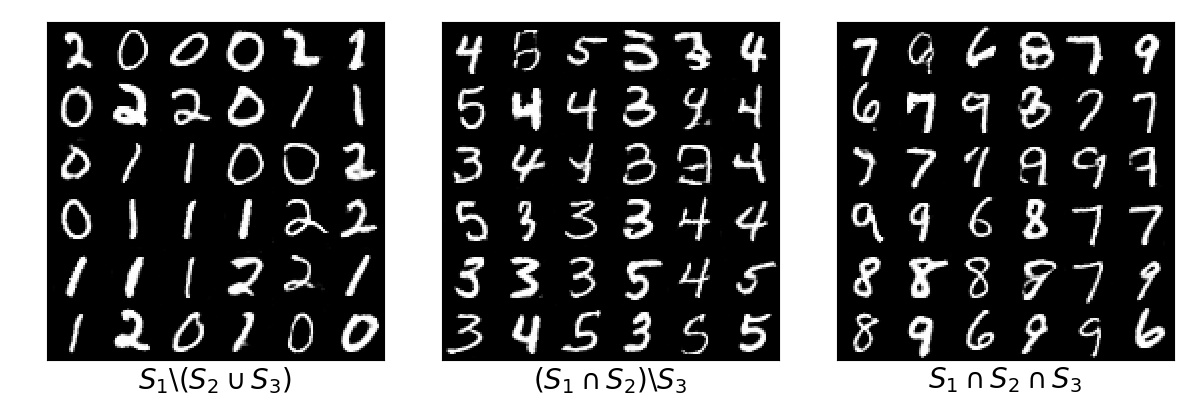}}
    \subfigure{\includegraphics[width=1.0\linewidth]{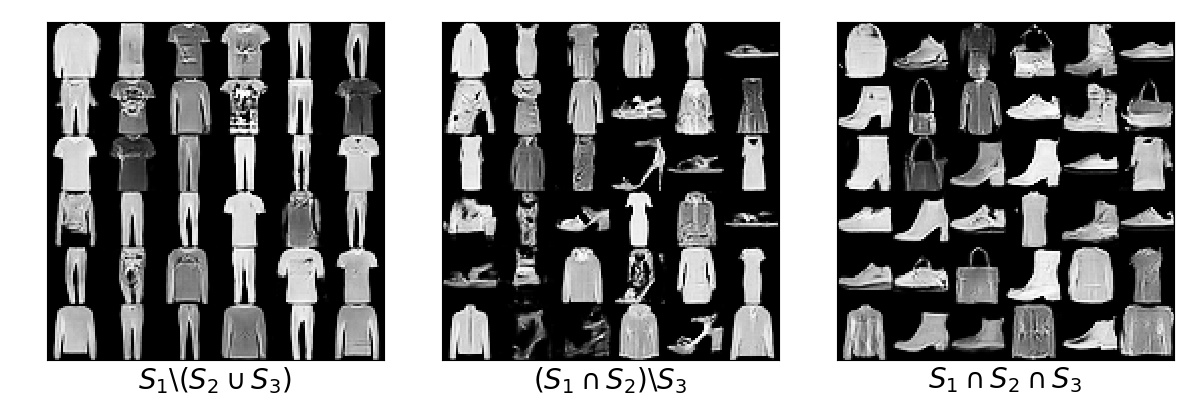}}
    \subfigure{\includegraphics[width=1.0\linewidth]{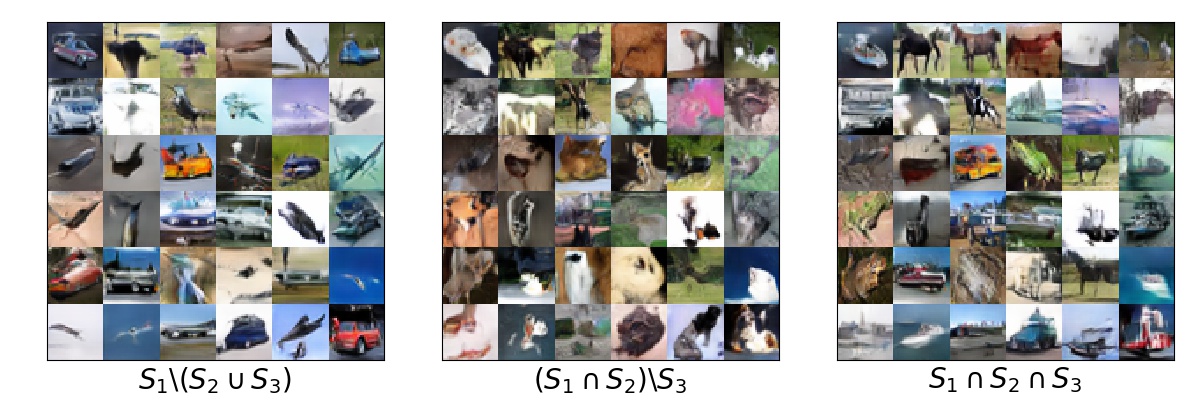}}
    \vspace{-4mm}
	\caption{MNIST, fashion-MNIST, CIFAR-10 results for case C}
	\label{case3}
\end{figure}

\begin{figure*}[ht!]
	\centering
    \subfigure{\includegraphics[width=1.0\linewidth]{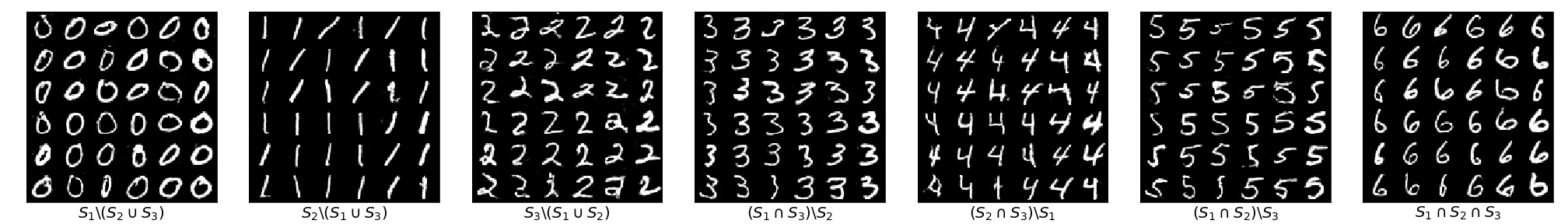}}
    \subfigure{\includegraphics[width=1.0\linewidth]{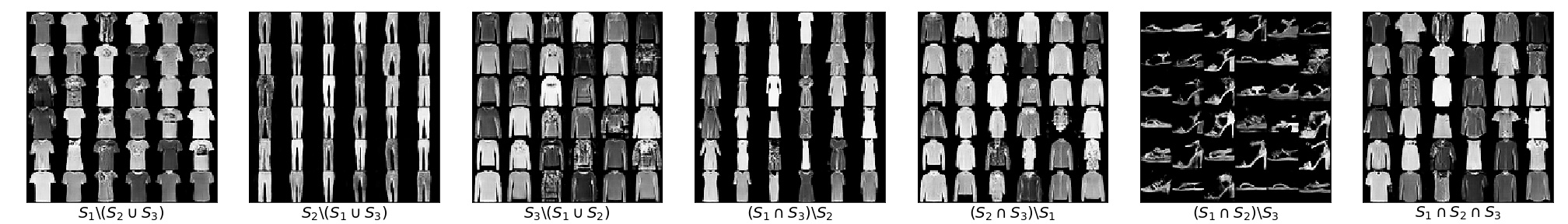}}
    \subfigure{\includegraphics[width=1.0\linewidth]{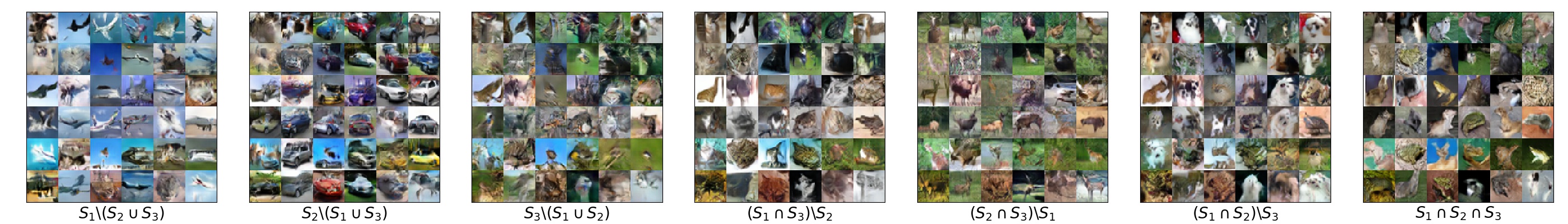}}
    \vspace{-4mm}
	\caption{MNIST, fashion-MNIST, CIFAR-10 results for case B}
	\label{case2}
\end{figure*}

Figure \ref{case1}, \ref{case2}, \ref{case3} shows the results for cases A, B, and C respectively. In case A of MNIST, $S_1 \setminus S_2 = \left\{ {0,1,2} \right\} $, $S_2 \setminus S_1 = \left\{ {7,8,9} \right\}$ and $S_2 \cap S_1 = \left\{ {3,4,5,6} \right\} $ are correctly modeled. Similarly, for Fashion-MNIST,  $S_1 \setminus S_2 = \left\{ {\textit{T-shirt/top},\textit{Trouser}, \textit{Pullover}} \right\} $, $S_2 \setminus S_1 = \left\{ {\textit{Sneaker}, \textit{Bag}, \textit{Ankle boot}} \right\}$ and $S_2 \cap S_1 = \left\{ {\textit{Dress}, \textit{Coat}, \textit{Sandal}, \textit{Shirt}} \right\} $ are correctly modeled. CIFAR-10 image quality is not as good as the others, so it is not easy to make a judgment. However, from the recognizable classes we see that ``Automobile'', ``Horse'', ``Ship'', ``Truck'' appears in the right region. In case B of MNIST and Fashion-MNIST, the vast majority of the object appears in the right region with good image quality. For CIFAR-10, the results are decent for ``Airplane'', ``Automobile'', ``Deer''. For other regions the quality is not satisfactory and there seems to be some amount of leaking. In case C, we see near perfect performance in case of MNIST and fashion-MNIST. Objects are placed in the right regions and image quality is good enough to recognize the objects. Image quality of CIFAR-10 is again not very good, but the objects seems to be placed in the right regions. For example $S_1 \setminus(S_2 \cap S_3)$ only includes ``Airplane'', ``Automobile'' and ``Bird'', while $S_1 \cup S_2 \cup S_3$ only includes ``Frog'', ``Horse'', ``Ship'' and ``Truck''.

\begin{table*}[ht!]
\caption{Quantitative results on 3 datasets and 3 cases. Accuracy of each region is reported. IG stands for Independent Generators and Avg is average of all the regions. The regions can be tracked from Figure~\ref{venn_diagrams}. n/a is placed into the regions where it does not exist in the type of Venn diagram.}
\centering
\small
\begin{tabular}{cccccccccccc}
 \hline
 Dataset & Case  & Classifier & IG & $r_1$ & $r_2$ & $r_3$ & $r_4$ & $r_5$ & $r_6$ & $r_7$ & Avg \\
 \hline
 MNIST &  A & Yes & Yes & 99.76 & 99.11 & 81.72 & n/a & n/a & n/a & n/a & 93.53\\
 MNIST &  A & Yes & No & 99.69 & 98.86 & 83.40 & n/a & n/a & n/a & n/a & 93.98\\
 F-MNIST &  A & Yes & Yes & 91.37 & 87.75 & 80.17 & n/a & n/a & n/a & n/a & 86.43\\
 F-MNIST &  A & Yes & No & 90.15 & 86.48 & 80.92 & n/a & n/a & n/a & n/a & 85.85\\
 CIFAR-10 &  A & Yes & Yes & 78.03 & 75.19 & 58.07 & n/a & n/a & n/a & n/a & 70.42\\
 CIFAR-10 &  A & Yes & No & 72.23 & 71.65 & 52.78 & n/a & n/a & n/a & n/a & 65.55\\
 \hline
 MNIST &  B & Yes & Yes & 99.33 & 100.0 & 95.67 & 98.44 & 98.22 & 99.64 & 99.58 & 98.70\\
 MNIST &  B & Yes & No & 99.32 & 100.0 & 96.05 & 98.75 & 98.14 & 99.56 & 99.36 & 98.74 \\
 F-MNIST &  B & Yes & Yes & 73.03 & 97.36 & 70.43 & 68.33 & 91.02 & 92.09 & 18.59 & 72.97\\
 F-MNIST &  B & Yes & No & 71.86 & 98.07 & 68.45 & 71.04 & 93.17 & 91.54 & 18.02 & 73.16\\
 CIFAR-10 &  B & Yes & Yes & 83.57 & 58.71 & 10.63 & 53.14 & 2.81 & 51.93 & 28.28 & 41.30\\
 CIFAR-10 &  B & Yes & No & 88.3 & 52.84 & 11.43 & 51.99 & 2.98 & 52.78 & 35.29 & 42.23 \\
 \hline
 MNIST &  C & Yes & Yes & 99.5 & n/a & n/a & n/a & n/a & 93.85 & 94.19 & 95.85\\
 MNIST &  C & Yes & No & 99.12 & n/a & n/a & n/a & n/a & 93.08 & 93.64 & 95.28\\
 F-MNIST &  C & Yes & Yes & 94.88 & n/a & n/a & n/a & n/a & 85.25 & 67.49 & 82.54\\
 F-MNIST &  C & Yes & No & 94.41 & n/a & n/a & n/a & n/a & 83.17 & 67.5 & 81.69\\
 CIFAR-10 &  C & Yes & Yes & 85.63 & n/a & n/a & n/a & n/a & 70.57 & 63.83 & 73.34\\
 CIFAR-10 &  C & Yes & No & 77.39 & n/a & n/a & n/a & n/a & 66.82 & 61.85 & 68.69\\
 \hline
 \hline
 MNIST &  A & No & No & 99.54 & 98.88 & 81.45 & n/a & n/a & n/a & n/a & 93.29 \\
 F-MNIST &  A & No & No & 90.48 & 86.59 & 80.12 & n/a & n/a & n/a & n/a & 85.73 \\
 CIFAR-10 &  A & No & Yes & 76.4 & 73.32 & 60.93 & n/a & n/a & n/a & n/a & 70.22 \\
 MNIST &  B & No & No & 98.72 & 99.99 & 95.28 & 99.08 & 97.40 & 99.29 & 99.29 & 98.43 \\
 F-MNIST &  B & No & No & 67.89 & 97.81 & 63.79 & 68.18 & 88.98 & 91.91 & 15.68 & 70.61 \\
 CIFAR-10 &  B &  No & Yes & 85.17 & 51.64 & 9.27 & 51.44 & 2.89 & 46.48 & 22.97 & 38.55\\ 
 MNIST &  C & No & No & 98.49 & n/a & n/a & n/a & n/a & 92.88 & 93.71 & 95.03\\
 F-MNIST &  C & No & No & 92.57 & n/a & n/a & n/a & n/a & 84.04 & 67.24 & 81.28\\
 CIFAR-10 &  C & No & Yes & 86.14 & n/a & n/a & n/a & n/a & 71.85 & 61.77 & 73.25\\
 \hline
\end{tabular}
\label{quantitative_results}
\end{table*}

Table~\ref{quantitative_results} lists quantitative results for the experiments above. Interestingly, MNIST performs best in case B, while the same case is the hardest for Fashion-MNIST and CIFAR-10. We believe this is due to the clear separation between the classes in MNIST, while there are a few hard to distinguish classes in Fashion-MNIST such as ``Pullover'', ``Coat'', ``Shirt''. As expected, average accuracy drops as the dataset becomes harder ($ \textrm{Acc(MNIST)}> \textrm{Acc(Fashion-MNIST)} > \textrm{Acc(CIFAR-10)}$).

\textbf{Conditional Generator vs.\ Independent Generators}: In case of MNIST and Fashion-MNIST, conditional generator produces comparable or slightly better results, while independent generators are better for CIFAR-10. We postulate that in case of simple datasets, single conditional generator has sufficient capacity to match the quality of multiple generators. Besides, sharing most of the weights with different regions regularizes the training, as there are many common features between regions. However when it comes to CIFAR-10, sharing weights might be a burden for the representation of different regions rather than a regularization.  

\textbf{Effect of the Classifier}: As explained in the method section, we have utilized a classifier to alleviate leaking issues between regions. In this section we evaluate its effectiveness on various datasets. In order to reduce the number of setting we use conditional generators for MNIST and Fashion-MNIST and independent generators for CIFAR-10 due to reasons explained in the previous section. The bottom section of Table~\ref{quantitative_results} belongs to 9 different settings without classifier term in the objective. At all settings there are slight but consistent improvements. For CIFAR-10, improvements are more significant than for the other datasets.


\begin{figure*}[ht!]
	\centering
    \includegraphics[width=1.0\linewidth]{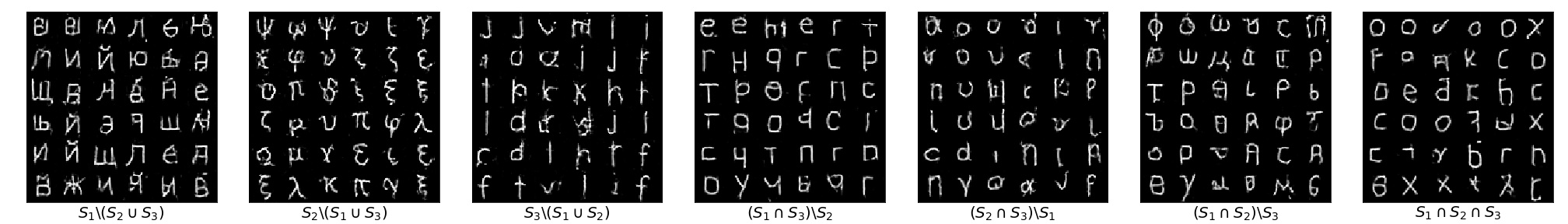}
    \vspace{-4mm}
	\caption{Omniglot results}
	\label{fig_omniglot}
\end{figure*}

\textbf{Omniglot} \cite{omniglot} contains letters from many alphabets. Each alphabet contains a certain number of letters, and there are $20$ samples per letter, which make this dataset hard to model. We have selected ``Cyrillic", ``Greek" and ``Latin" alphabets as 3 different distributions. As these alphabets include both unique and common letters, we aim to model it with $d_2$ type modeling to discover both unique and common letters. 

In Figure \ref{fig_omniglot}, the first three regions corresponds to only ``Cyrillic'', ``Greek'' and ``Latin'' in order. The majority of the letters in each  of these regions belongs to their own alphabet and not in others. For other regions there are more mistakes like the letter ``o'' appearing in multiple regions.

\textbf{CelebA} \cite{celeba}: For this dataset, we use both $d_1$ and $d_3$ types with two distributions. In case of $d_1$, the first distribution contains only male faces while the second one contains females. In case of $d_3$, the first distribution contains only female faces while the second one contains both genders. Our aim is to see whether semantic commonalities and differences of the distributions can be captured successfully. In $d_1$ setting, there should be no overlap in genders but we are interested in what type of commonalities our method can find. We have used conditional generator for this experiment to see the  semantic relations between the regions more clearly.

\begin{figure*}[ht!]
	\centering
    \includegraphics[width=1.0\linewidth]{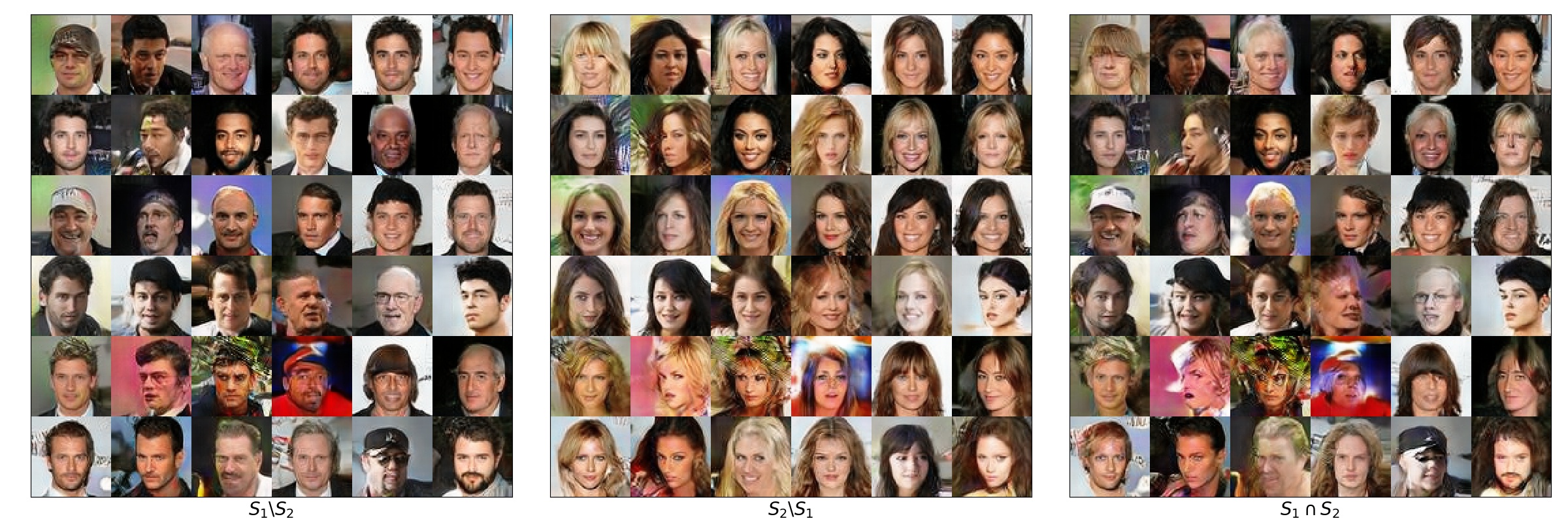}
    \vspace{-4mm}
	\caption{CelebA results: $S_1 \setminus S_2$ is only males, $S_2 \setminus S_1$ is only females, $S_1 \cap S_2$ is intersection}
	\label{fig_celeba_v1}
\end{figure*}

\begin{figure}[ht!]
	\centering
    \includegraphics[width=1.0\linewidth]{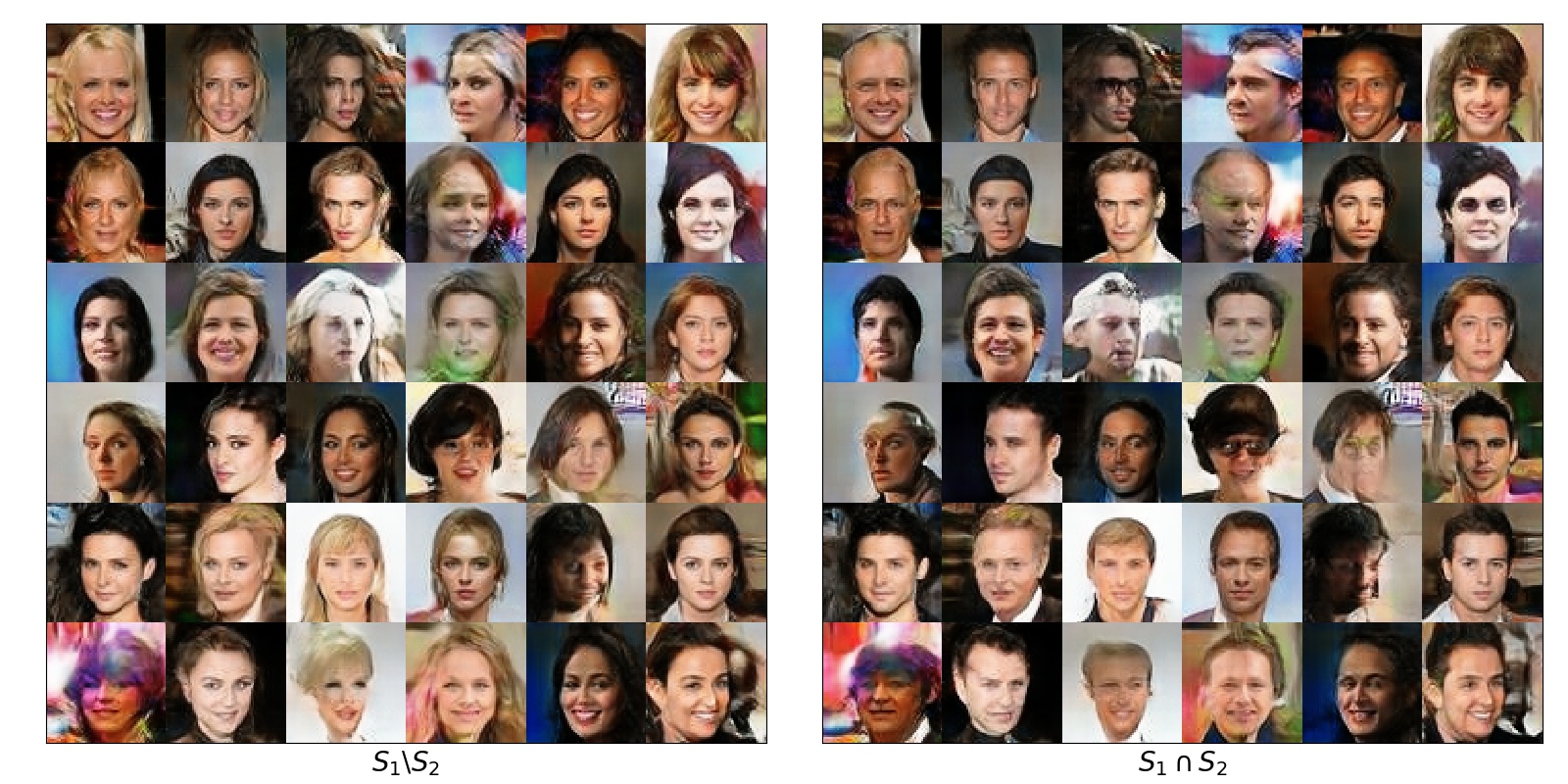}
    \vspace{-4mm}
	\caption{CelebA results: $S_1 \setminus S_2$ is only females, $S_1 \cap S_2$ is only males}
	\label{fig_celeba_v3}
\end{figure}

In CelebA $d_1$ (Figure \ref{fig_celeba_v1}), $S_1 \setminus S_2$ depicts stereotype masculine faces with short hair and masculine faces, whereas $S_2 \setminus S_1$ exhibits predominantly feminine features like long hair etc. On the other hand, $S_1 \cap S_2$ features faces which are neither predominantly male nor female. As the images in different regions are generated with the same noise, pose and background of an image at different regions remain similar, while the facial attributes change. Similarly, CelebA $d_3$ (Figure \ref{fig_celeba_v3}) shows that the model can capture commonalities of the distributions well, $S_1 \cap S_2$, correctly with all male faces, while the difference, $S_1 \setminus S_2$, are female faces as it should be. Again, due to the same noise, generations between different regions can be compared. Both experiments show that Venn~GAN can capture high level semantic commonality between high dimensional complex distributions.

\section{Discussion \& Conclusion}
In this paper, we have used prior knowledge to choose the Venn type or $\boldsymbol{O}$ matrix. When we know that the distributions have intersections and unique parts, $d_1$ or $d_2$ type has been used; if a distribution is subset of another one, then we have utilized $d_3$. We note that certain distributions may not fall under either one of those two types. If we have a prior knowledge about the type of the distributions, then this method can be utilized easily. In case we have no prior knowledge about it, the ideal situation would be learning it, which we leave for future work.

The main limitation of the method is that it takes union over each region with equal probability, which is a strong assumption in many cases. In an ideal situation we should optimize $\boldsymbol{O}$ end-to-end with the model parameters. One challenge is that the mixture weights are discrete, as in practice we use the number of samples to approximate them. However this can be handled with a reinforcement learning algorithm. Another bigger challenge is to find a meaningful reward signal for the training of $\boldsymbol{O}$. This reward should negatively correlate with ``leaks'' between the regions. We think this is also an important future research direction.

In conclusion, we have proposed a novel multi-distribution GAN method which can discover particularities and commonalities between distributions. Our method models each data distribution with a mixture of generator distributions. As the generators are partially shared between the modeling of different true data distributions, shared ones captures the commonality of the distributions, while non-shared ones capture unique aspects of them. We have successfully trained it on various datasets to show its effectiveness. We believe this method has good potential for new applications and better data modeling.

\subsubsection*{Acknowledgments}
Yasin Yaz{\i}c{\i} was supported by a SINGA scholarship from the Agency for Science, Technology and Research (A*STAR). Georgios Piliouras would like to acknowledge SUTD grant SRG ESD 2015 097, MOE AcRF Tier 2 Grant 2016-T2-1-170 and NRF 2018 Fellowship NRF-NRFF2018-07. This research is partially supported by the Agency for Science, Technology and Research (A*STAR) under its AME Programmatic Funds (Project No.A1892b0026). This research was carried out at Advanced Digital Sciences Center (ADSC), Institute for Infocomm Research (I2R) and at the Rapid-Rich Object Search (ROSE) Lab at the Nanyang Technological University, Singapore. The ROSE Lab is supported by the National Research Foundation, Singapore, and the Infocomm Media Development Authority, Singapore. Research at I2R was partially supported by A*STAR SERC Strategic Funding (A1718g0045). The computational work for this article was partially performed on resources of the National Supercomputing Centre, Singapore (https://www.nscc.sg).

\bibliography{ms}

\begin{thebibliography}{31}
\providecommand{\natexlab}[1]{#1}
\providecommand{\url}[1]{\texttt{#1}}
\expandafter\ifx\csname urlstyle\endcsname\relax
  \providecommand{\doi}[1]{doi: #1}\else
  \providecommand{\doi}{doi: \begingroup \urlstyle{rm}\Url}\fi

\bibitem[Arora et~al.(2017)Arora, Ge, Liang, Ma, and
  Zhang]{DBLP:journals/corr/Arora0LMZ17}
Arora, S., Ge, R., Liang, Y., Ma, T., and Zhang, Y.
\newblock Generalization and equilibrium in generative adversarial nets
  {(GANs)}.
\newblock \emph{CoRR}, abs/1703.00573, 2017.
\newblock URL \url{http://arxiv.org/abs/1703.00573}.

\bibitem[de~Vries et~al.(2017)de~Vries, Strub, Mary, Larochelle, Pietquin, and
  Courville]{devries+al-2017-modulating}
de~Vries, H., Strub, F., Mary, J., Larochelle, H., Pietquin, O., and Courville,
  A.
\newblock Modulating early visual processing by language.
\newblock In \emph{Advances in Neural Information Processing Systems 30 (NIPS
  2017)}, pp.\  6597--6607, December 2017.
\newblock arxiv: 1707.00683.

\bibitem[{Doan} et~al.(2018){Doan}, {Monteiro}, {Albuquerque}, {Mazoure},
  {Durand}, {Pineau}, and {Devon Hjelm}]{gan_multi_armed_bandit}
{Doan}, T., {Monteiro}, J., {Albuquerque}, I., {Mazoure}, B., {Durand}, A.,
  {Pineau}, J., and {Devon Hjelm}, R.
\newblock On-line adaptative curriculum learning for {GANs}.
\newblock \emph{ArXiv e-prints}, July 2018.

\bibitem[Dumoulin et~al.(2017)Dumoulin, Shlens, and
  Kudlur]{dumoulin2017learned-iclr}
Dumoulin, V., Shlens, J., and Kudlur, M.
\newblock A learned representation for artistic style.
\newblock In \emph{International Conference on Learning Representations 2017
  (Conference Track)}, 2017.
\newblock URL \url{https://openreview.net/forum?id=BJO-BuT1g}.

\bibitem[Durugkar et~al.(2016)Durugkar, Gemp, and Mahadevan]{multi_disc_ishan}
Durugkar, I.~P., Gemp, I., and Mahadevan, S.
\newblock Generative multi-adversarial networks.
\newblock \emph{CoRR}, abs/1611.01673, 2016.
\newblock URL \url{http://arxiv.org/abs/1611.01673}.

\bibitem[Ghosh et~al.(2017)Ghosh, Kulharia, Namboodiri, Torr, and
  Dokania]{madgan}
Ghosh, A., Kulharia, V., Namboodiri, V.~P., Torr, P. H.~S., and Dokania, P.~K.
\newblock Multi-agent diverse generative adversarial networks.
\newblock \emph{CoRR}, abs/1704.02906, 2017.
\newblock URL \url{http://arxiv.org/abs/1704.02906}.

\bibitem[Goodfellow et~al.(2014)Goodfellow, Pouget-Abadie, Mirza, Xu,
  Warde-Farley, Ozair, Courville, and Bengio]{gan}
Goodfellow, I., Pouget-Abadie, J., Mirza, M., Xu, B., Warde-Farley, D., Ozair,
  S., Courville, A., and Bengio, Y.
\newblock Generative adversarial nets.
\newblock In \emph{Advances in Neural Information Processing Systems 27}, pp.\
  2672--2680. 2014.
\newblock URL
  \url{http://papers.nips.cc/paper/5423-generative-adversarial-nets.pdf}.

\bibitem[Gulrajani et~al.(2017)Gulrajani, Ahmed, Arjovsky, Dumoulin, and
  Courville]{wgan_gp}
Gulrajani, I., Ahmed, F., Arjovsky, M., Dumoulin, V., and Courville, A.
\newblock Improved training of {Wasserstein GANs}.
\newblock pp.\  5769--5779, December 2017.
\newblock arxiv: 1704.00028.

\bibitem[Hoang et~al.(2017)Hoang, Nguyen, Le, and Phung]{mggan}
Hoang, Q., Nguyen, T.~D., Le, T., and Phung, D.~Q.
\newblock Multi-generator generative adversarial nets.
\newblock \emph{CoRR}, abs/1708.02556, 2017.
\newblock URL \url{http://arxiv.org/abs/1708.02556}.

\bibitem[Hoang et~al.(2018)Hoang, Nguyen, Le, and Phung]{hoang2018mgan}
Hoang, Q., Nguyen, T.~D., Le, T., and Phung, D.
\newblock {MGAN}: Training generative adversarial nets with multiple
  generators.
\newblock In \emph{International Conference on Learning Representations}, 2018.
\newblock URL \url{https://openreview.net/forum?id=rkmu5b0a-}.

\bibitem[Isola et~al.(2016)Isola, Zhu, Zhou, and
  Efros]{DBLP:journals/corr/IsolaZZE16}
Isola, P., Zhu, J., Zhou, T., and Efros, A.~A.
\newblock Image-to-image translation with conditional adversarial networks.
\newblock \emph{CoRR}, abs/1611.07004, 2016.
\newblock URL \url{http://arxiv.org/abs/1611.07004}.

\bibitem[Juefei{-}Xu et~al.(2017)Juefei{-}Xu, Boddeti, and
  Savvides]{DBLP:journals/corr/Juefei-XuBS17}
Juefei{-}Xu, F., Boddeti, V.~N., and Savvides, M.
\newblock Gang of gans: Generative adversarial networks with maximum margin
  ranking.
\newblock \emph{CoRR}, abs/1704.04865, 2017.
\newblock URL \url{http://arxiv.org/abs/1704.04865}.

\bibitem[Kaneko et~al.(2018)Kaneko, Ushiku, and
  Harada]{DBLP:journals/corr/abs-1811-11163}
Kaneko, T., Ushiku, Y., and Harada, T.
\newblock Class-distinct and class-mutual image generation with gans.
\newblock \emph{CoRR}, abs/1811.11163, 2018.
\newblock URL \url{http://arxiv.org/abs/1811.11163}.

\bibitem[Karras et~al.(2017)Karras, Aila, Laine, and Lehtinen]{pggan}
Karras, T., Aila, T., Laine, S., and Lehtinen, J.
\newblock Progressive growing of gans for improved quality, stability, and
  variation.
\newblock \emph{CoRR}, abs/1710.10196, 2017.
\newblock URL \url{http://arxiv.org/abs/1710.10196}.

\bibitem[Kingma \& Ba(2014)Kingma and Ba]{adam}
Kingma, D.~P. and Ba, J.
\newblock Adam: {A} method for stochastic optimization.
\newblock \emph{CoRR}, abs/1412.6980, 2014.
\newblock URL \url{http://arxiv.org/abs/1412.6980}.

\bibitem[Krizhevsky et~al.()Krizhevsky, Nair, and Hinton]{cifar10}
Krizhevsky, A., Nair, V., and Hinton, G.
\newblock Cifar-10 (canadian institute for advanced research).
\newblock URL \url{http://www.cs.toronto.edu/~kriz/cifar.html}.

\bibitem[Lake et~al.(2015)Lake, Salakhutdinov, and Tenenbaum]{omniglot}
Lake, B.~M., Salakhutdinov, R., and Tenenbaum, J.~B.
\newblock Human-level concept learning through probabilistic program induction.
\newblock \emph{Science}, 350\penalty0 (6266):\penalty0 1332--1338, 2015.
\newblock \doi{10.1126/science.aab3050}.

\bibitem[Ledig et~al.(2016)Ledig, Theis, Huszar, Caballero, Aitken, Tejani,
  Totz, Wang, and Shi]{DBLP:journals/corr/LedigTHCATTWS16}
Ledig, C., Theis, L., Huszar, F., Caballero, J., Aitken, A.~P., Tejani, A.,
  Totz, J., Wang, Z., and Shi, W.
\newblock Photo-realistic single image super-resolution using a generative
  adversarial network.
\newblock \emph{CoRR}, abs/1609.04802, 2016.
\newblock URL \url{http://arxiv.org/abs/1609.04802}.

\bibitem[Liu et~al.(2015)Liu, Luo, Wang, and Tang]{celeba}
Liu, Z., Luo, P., Wang, X., and Tang, X.
\newblock Deep learning face attributes in the wild.
\newblock In \emph{Proc. International Conference on Computer Vision (ICCV)},
  2015.

\bibitem[Mescheder(2018)]{convergence_lars}
Mescheder, L.~M.
\newblock On the convergence properties of {GAN} training.
\newblock \emph{CoRR}, abs/1801.04406, 2018.
\newblock URL \url{http://arxiv.org/abs/1801.04406}.

\bibitem[Mirza \& Osindero(2014)Mirza and
  Osindero]{DBLP:journals/corr/MirzaO14}
Mirza, M. and Osindero, S.
\newblock Conditional generative adversarial nets.
\newblock \emph{CoRR}, abs/1411.1784, 2014.
\newblock URL \url{http://arxiv.org/abs/1411.1784}.

\bibitem[Miyato \& Koyama(2018)Miyato and Koyama]{miyato2018cgans}
Miyato, T. and Koyama, M.
\newblock c{GAN}s with projection discriminator.
\newblock In \emph{International Conference on Learning Representations}, 2018.
\newblock URL \url{https://openreview.net/forum?id=ByS1VpgRZ}.

\bibitem[Miyato et~al.(2018)Miyato, Kataoka, Koyama, and
  Yoshida]{spectral_norm}
Miyato, T., Kataoka, T., Koyama, M., and Yoshida, Y.
\newblock Spectral normalization for generative adversarial networks.
\newblock \emph{CoRR}, abs/1802.05957, 2018.
\newblock URL \url{http://arxiv.org/abs/1802.05957}.

\bibitem[Neyshabur et~al.(2017)Neyshabur, Bhojanapalli, and
  Chakrabarti]{DBLP:journals/corr/NeyshaburBC17}
Neyshabur, B., Bhojanapalli, S., and Chakrabarti, A.
\newblock Stabilizing {GAN} training with multiple random projections.
\newblock \emph{CoRR}, abs/1705.07831, 2017.
\newblock URL \url{http://arxiv.org/abs/1705.07831}.

\bibitem[{Odena} et~al.(2016){Odena}, {Olah}, and
  {Shlens}]{2016arXiv161009585O}
{Odena}, A., {Olah}, C., and {Shlens}, J.
\newblock {Conditional Image Synthesis With Auxiliary Classifier GANs}.
\newblock \emph{ArXiv e-prints}, October 2016.

\bibitem[{Radford} et~al.(2015){Radford}, {Metz}, and {Chintala}]{dcgan}
{Radford}, A., {Metz}, L., and {Chintala}, S.
\newblock {Unsupervised Representation Learning with Deep Convolutional
  Generative Adversarial Networks}.
\newblock \emph{CoRR}, abs/1511.06434, 2015.
\newblock URL \url{https://arxiv.org/abs/1511.06434}.

\bibitem[Reed et~al.(2016)Reed, Akata, Yan, Logeswaran, Schiele, and
  Lee]{DBLP:journals/corr/ReedAYLSL16}
Reed, S.~E., Akata, Z., Yan, X., Logeswaran, L., Schiele, B., and Lee, H.
\newblock Generative adversarial text to image synthesis.
\newblock \emph{CoRR}, abs/1605.05396, 2016.
\newblock URL \url{http://arxiv.org/abs/1605.05396}.

\bibitem[{Tolstikhin} et~al.(2017){Tolstikhin}, {Gelly}, {Bousquet},
  {Simon-Gabriel}, and {Sch{\"o}lkopf}]{adaGAN}
{Tolstikhin}, I., {Gelly}, S., {Bousquet}, O., {Simon-Gabriel}, C.-J., and
  {Sch{\"o}lkopf}, B.
\newblock {AdaGAN}: Boosting generative models.
\newblock \emph{ArXiv e-prints}, January 2017.

\bibitem[Xiao et~al.(2017)Xiao, Rasul, and Vollgraf]{fashion_mnist}
Xiao, H., Rasul, K., and Vollgraf, R.
\newblock Fashion-{MNIST}: a novel image dataset for benchmarking machine
  learning algorithms.
\newblock \emph{CoRR}, abs/1708.07747, 2017.
\newblock URL \url{https://arxiv.org/abs/1708.07747}.

\bibitem[{Yaz{\i}c{\i}} et~al.(2018){Yaz{\i}c{\i}}, {Foo}, {Winkler}, {Yap},
  {Piliouras}, and {Chandrasekhar}]{yasin_ema}
{Yaz{\i}c{\i}}, Y., {Foo}, C.-S., {Winkler}, S., {Yap}, K.-H., {Piliouras}, G.,
  and {Chandrasekhar}, V.
\newblock The unusual effectiveness of averaging in {GAN} training.
\newblock \emph{CoRR}, abs/1806.04498, June 2018.
\newblock URL \url{https://arxiv.org/abs/1806.04498}.

\bibitem[Yi et~al.(2017)Yi, Zhang, Tan, and Gong]{DBLP:journals/corr/YiZTG17}
Yi, Z., Zhang, H., Tan, P., and Gong, M.
\newblock Dualgan: Unsupervised dual learning for image-to-image translation.
\newblock \emph{CoRR}, abs/1704.02510, 2017.
\newblock URL \url{http://arxiv.org/abs/1704.02510}.

\end{thebibliography}
\bibliographystyle{icml2019}

\newpage
\appendix


\section{Network Architectures}
Prior distribution for the generator(s) is a 128-dimensional isotropic Gaussian distribution. If not mentioned, stride and padding of the convolution is $1$. ``Cond'' is conditioning which is linear scaling and addition for each feature channel. It is not used when multiple generators utilized. ``LReLU'' is LeakyReLU with $\alpha=0.2$.

\begin{table}[ht!]
\caption{Generator Architecture for 28x28 resolution (MNIST, Fashion-MNIST, Omniglot)}
\begin{center}
\begin{tabular}{ lcc } 
 \toprule
 Layers & Act. & Output Shape \\
 \midrule
 Latent vector & - & 128 x 1 x 1 \\
 Conv 4 x 4, pad=3 & Cond - LReLU & 128 x 4 x 4 \\
 Conv 4 x 4, pad=3 & Cond - LReLU & 128 x 7 x 7 \\
 Upsample & - & 128 x 14 x 14 \\
 Conv 3 x 3, pad=1 & Cond - LReLU & 64 x 14 x 14 \\
 Upsample & - & 64 x 28 x 28 \\
 Conv 3 x 3, pad=1 & Cond - LReLU & 32 x 28 x 28 \\
 Conv 3 x 3, pad=1 & Tanh & 1 x 28 x 28 \\
 \bottomrule
\end{tabular}
\end{center}
\end{table}

\begin{table}[ht!]
\caption{Discriminator Architecture for 28x28 resolution (MNIST, Fashion-MNIST, Omniglot)}
\begin{center}
\begin{tabular}{ lcc } 
 \toprule
 Layers & Act. & Output Shape \\
 \midrule
 Input image & - & 3 x 28 x 28 \\
 Conv 4 x 4, st=3 & LReLU & 64 x 14 x 14 \\
 Conv 4 x 4, st=3 & LReLU & 128 x 7 x 7 \\
 Conv 4 x 4, st=3 & LReLU & 256 x 3 x 3 \\
 Conv 3 x 3, st=1, pad=0 & Squeeze & 1 \\
 \bottomrule
\end{tabular}
\end{center}
\end{table}

\begin{table}[ht!]
\caption{Generator Architecture for 32x32 resolution (CIFAR-10)}
\begin{center}
\begin{tabular}{ lcc } 
 \toprule
 Layers & Act. & Output Shape \\
 \midrule
 Latent vector & - & 128 x 1 x 1 \\
 Conv 4 x 4, pad=3 & Cond - LReLU & 512 x 4 x 4 \\
 Upsample & - & 512 x 8 x 8  \\
 Conv 3 x 3 & Cond - LReLU & 256 x 8 x 8 \\
 Upsample & - & 256 x 16 x 16  \\
 Conv 3 x 3 & Cond - LReLU & 128 x 16 x 16 \\
 Upsample & - & 128 x 32 x 32 \\
 Conv 3 x 3 & Cond - LReLU & 64 x 32 x 32 \\
 Conv 3 x 3 & Tanh & 3 x 32 x 32 \\
 \bottomrule
\end{tabular}
\end{center}
\end{table}

\begin{table}[ht!]
\caption{Discriminator Architecture for 32x32 resolution (CIFAR-10)}
\begin{center}
\begin{tabular}{ lcc } 
 \toprule
 Layers & Act. & Output Shape \\
 \midrule
 Input image & - & 3 x 32 x 32 \\
 Conv 3 x 3 & LReLU & 64 x 32 x 32 \\
 Conv 3 x 3 & LReLU & 128 x 32 x 32 \\
 Downsample & - & 128 x 16 x 16 \\
 \hline
 Conv 3 x 3 & LReLU & 128 x 16 x 16 \\
 Conv 3 x 3 & LReLU & 256 x 16 x 16 \\
 Downsample & - & 256 x 8 x 8 \\
 \hline
 Conv 3 x 3 & LReLU & 256 x 8 x 8 \\
 Conv 3 x 3 & LReLU & 512 x 8 x 8 \\
 Downsample & - & 512 x 4 x 4 \\
 \hline
 Conv 4 x 4, st=1, pad=0 & Squeeze & 1 \\
 \bottomrule
\end{tabular}
\end{center}
\end{table}

\begin{table}[ht!]
\caption{ResNet Generator Architecture for 64x64 resolution (CelebA)}
\begin{center}
\begin{tabular}{ lcc } 
 \toprule
 Layers & Act. & Output Shape \\
 \midrule
 Latent vector & - & 128 x 1 x 1 \\
 Conv 4 x 4, pad=3 & Cond & 512 x 4 x 4 \\
 ResBlock  & - & 512 x 4 x 4 \\
 Upsample & Cond & 512 x 8 x 8  \\
 \hline
 ResBlock  & - & 512 x 8 x 8 \\
 Upsample & Cond & 512 x 16 x 16  \\
 \hline
 ResBlock  & - & 256 x 16 x 16 \\
 Upsample & Cond & 256 x 32 x 32  \\
 \hline
 ResBlock  & - & 128 x 32 x 32 \\
 Upsample & Cond & 128 x 64 x 64  \\
 \hline
 ResBlock  & LReLU - Cond & 64 x 64 x 64 \\
 Conv 3 x 3 & Tanh & 3 x 64 x 64  \\
 \bottomrule
\end{tabular}
\end{center}
\end{table}

\begin{table}[H]
\caption{ResNet Discriminator Architecture for 64x64 resolution (CelebA)}
\begin{center}
\begin{tabular}{ lcc } 
 \toprule
 Layers & Act. & Output Shape \\
 \midrule
 Input image & - & 3 x 64 x 64 \\
 Conv 3 x 3 & - & 64 x 64 x 64 \\
 ResBlock  & - & 64 x 64 x 64 \\
 Downsample & - & 64 x 32 x 32  \\
 \hline
 ResBlock  & - & 128 x 32 x 32 \\
 Downsample & - & 128 x 16 x 16  \\
 \hline
 ResBlock  & - & 256 x 16 x 16 \\
 Downsample & - & 256 x 8 x 8  \\
 \hline
 ResBlock  & - & 512 x 8 x 8 \\
 Downsample & - & 512 x 4 x 4  \\
 \hline
 ResBlock  & LReLU & 512 x 4 x 4 \\
 Conv 4 x 4, st=1, pad=0 & Squeeze & 1 \\
 \bottomrule
\end{tabular}
\end{center}
\end{table}

\section{Training of the classifiers for Quantification}
For MNIST, Fashion-MNIST and CIFAR-10, we have trained 3 separate classifier to assess quality of the method. For each dataset, the architecture is the same with the discriminator used for that dataset except the last layer which outputs $10$ logits value instead of $1$. We have used ADAM optimizer with learning rate of $0.0002$, $\beta_1 = 0.5$ and $\beta_2 =0.9$. Each model has been trained for 50k iterations with a batch size of $64$. The accuracy of the classifier on test sets for MNIST, fashion-MNIST and CIFAR-10 are 99.12, 91.20 and 84.20 respectively.

\section{Illustrative Examples}
For this experiments, we have used $7$ generators and $3$ discriminators. The network architecture for generators is 4 fully connected layer followed by LeakyReLU except the last one which is linear. The discriminators' are constructed from 4 fully connected layers followed by LeakyReLU except the last one which is linear. In both networks, each layer has 256 units while last layer of generator has $2$ and last layer of the discriminator has $1$. Prior distribution for the generators is a 128-dimensional isotropic Gaussian distribution. We have used ADAM \cite{adam} optimizer with learning rate of $0.0002$, $\beta_1 = 0.0$ and $\beta_2 =0.9$. The optimization of discriminator and generator follows alternating update rule with single discriminator update per generator update. The model has been trained for 5k iterations. For each region (generator), we use a batch size of $64$. $\lambda$ of $R_1$ regularizer is $0.1$.

\end{document}